# Adaptive Terminal Sliding Mode Control Using Deep Reinforcement Learning for Zero-Force Control of Exoskeleton Robot Systems

Morteza Mirzaee, Reza Kazemi

Faculty of Mechanical Engineering, K.N. Toosi University of Technology, Tehran, Iran.

**Abstract:**

This paper introduces a novel zero-force control method for upper-limb exoskeleton robots, which are used in a variety of applications including rehabilitation, assistance, and human physical capability enhancement. The proposed control method employs an Adaptive Integral Terminal Sliding Mode (AITSM) controller, combined with an exponential reaching law and Proximal Policy Optimization (PPO), a type of Deep Reinforcement Learning (DRL). The PPO system incorporates an attention mechanism and Long Short-Term Memory (LSTM) neural networks, enabling the controller to selectively focus on relevant system states, adapt to changing behavior, and capture long-term dependencies. This controller is designed to manage a 5-DOF upper-limb exoskeleton robot with zero force, even amidst system uncertainties. The controller uses an integral terminal sliding surface to ensure finite-time convergence to the desired state, a crucial feature for applications requiring quick responses. It also includes an exponential switching control term to reduce chattering and improve system accuracy. The controller's adaptability, facilitated by the PPO system, allows real-time parameter adjustments based on system feedback, making the controller robust and capable of dealing with uncertainties and disturbances that could affect the performance of the exoskeleton. The proposed control method's effectiveness and superiority are confirmed through numerical simulations and comparisons with existing control methods.

## 1. Introduction

Exoskeleton robotic systems, particularly upper-limb exoskeleton robots, are designed to enhance human locomotion and dexterity. They are used in various applications, including rehabilitation and power augmentation [1]. One of the key features of these robots is force compensation, which manages the forces that the robot applies to the user's limb, allowing tasks to be carried out with enhanced power and accuracy [2]. A specific force control strategy used is zero force control, which minimizes the force applied by the robot, making it "unnoticeable" to the user [3]. This is beneficial in situations where the user needs to move without any obstruction from the robot. The development of effective control technology to improve the practicability of exoskeleton robots is a major direction for future research in this field [4]. Furthermore, advancements in control systems will enable exoskeleton robots to adapt to individual users' needs and preferences, leading to more personalized and effective assistance.

Deep learning has had a profound impact on a wide range of distinct applications, from computer vision [5], [6] and natural language processing to speech recognition and healthcare, revolutionizing the way we approach tasks such as image classification, object detection, and predictive analytics [7], [8]. In recent years, the domain of control systems has gravitated towards the use of intelligent control and identification techniques, with a strong focus on machine learning and artificial intelligence [9]. Deep Reinforcement Learning (DRL), a method that learns from experience and navigates intricate environments, has surfaced as a promising instrument for managing industrial systems [10]. DRL is notably effective in stabilizing hyperchaotic systems that face uncertain parameters or disturbances. It is adept at handling high-dimensional state spaces and continuous action spaces, making it an ideal choice for controlling upper-limb rehabilitation exoskeleton systems [11]. [12] presents a DRL approach that enables soft robotic arms to learn pushing tasks, achieving robust and adaptive performance in various scenarios, including uncertain object dynamics and changing environmental conditions. [13] proposes a model tree-based approach for explaining the decision-making process of DRL agents in real-time robotic applications, providing interpretable and transparent insights into the agent's behavior and improving trust and understanding in human-robot collaboration. [14] presents a deep reinforcement learning approach that enables a bipedal robot to learn agile soccer skills, such as dribbling and shooting, through trial and error, achieving robust and dynamic movements in a simulated soccer environment.

Proximal Policy Optimization (PPO) is a model-free, on-policy deep reinforcement learning algorithm that has gained popularity in recent years [15]. PPO is designed to optimize policies in environments with high-dimensional state and action spaces [16]. It works by iteratively updating the policy parameters to maximize the expected cumulative reward, while also ensuring that the updated policy remains close to the previous policy. This proximity constraint helps to prevent large changes to the policy, which can lead to instability and poor performance. In [17], the focus is on designing a PPO-based attitude controller for a tilt-rotor UAV during the transition process. In [18], the authors propose a novel SMC adaptive actor-critic optimal control approach for switched nonlinear systems with an emphasis on average dwell time. [19] presents the design and development of a robotic ankle foot orthosis, powered by dielectric elastomer artificial muscle, for children with cerebral palsy. [20] presents a novel algebraic transformation approach to achieve consensus in multi-agent singular systems, providing a promising solution for distributed control and coordination of complex systems. [21] presents a reinforcement learning approach to learn continuous control actions for robotic grasping, demonstrating improved performance in a robotic arm grasping task. [22] compares the performance of PPO, and other reinforcement learning algorithms for generating walking gaits in quadruped robots, finding that PPO outperforms the others in terms of gait stability, speed, and robustness. [23] proposes a standing support arm design for robotic wheelchairs, utilizing a (PPO)-based reinforcement learning strategy to learn a control policy that assists users in standing up and sitting down safely and efficiently.

The attention mechanism is a technique used in deep learning models to allow the model to focus on specific parts of the input data that are relevant to the task at hand [24]. In the controller structure, the attention mechanism is used to selectively focus on specific parts of the system's state sequence, allowing the controller to weigh the importance of different elements and compute a context vector that captures the most relevant information. This enables the controller to adapt to changing system behavior and make more informed decisions [25]. LSTM networks are a type of recurrent neural network (RNN) that are designed to handle sequential data, such as time series data or natural language text. The LSTM network is used in the controller to capture long-term dependencies in the system's behavior, allowing it to learn patterns and relationships that span multiple time steps. This enables the controller to anticipate and respond to changes in the system, making it more effective at regulating the system's behavior. The LSTM network processes the system's state sequence and outputs a hidden state that is used to compute the control gains, which are then used to generate the control input [26]. [27] propose a DRL framework to optimize the control policy and attention mechanisms to focus on relevant system states. [28] presents a novel attention-LSTM model that incorporates a second-order learning algorithm to improve the accuracy and efficiency of dynamic chemical process modeling, enabling better prediction and control of complex chemical processes. [29] proposes a hydrological data prediction model that combines LSTM networks with an attention mechanism, enabling accurate and efficient forecasting of hydrological time series data, such as river flow and water level.

This paper presents an Adaptive Integral Terminal Sliding Mode (AITSM) controller, which incorporates an exponential reaching law and a PPO system with a deep learning network. This setup is utilized for zero-force control of a 5-DOF upper-limb exoskeleton robotic system under conditions with bounded uncertainties and external force disturbances. The key innovations of the proposed method are outlined as follows.

• **Adaptive Control**: The AITSM controller is designed to be adaptive, adjusting its reaching law parameters in real-time based on the PPO as an intelligent system. This adaptability enhances the controller’s resilience, enabling it to effectively handle uncertainties and external disturbances that could affect the performance of the exoskeleton.

• **Attention Mechanism**: The attention mechanism enables adaptive control systems to selectively focus on relevant system states, adapt to changing behavior, and capture long-term dependencies, leading to more efficient and effective control decisions. By prioritizing the most critical system states, the attention mechanism reduces the complexity of the control problem, allowing the controller to respond more quickly and accurately to changing system conditions.

• **LSTM Neural Networks**: The Captures long-term dependencies and enables the controller to make more informed decisions based on historical data. By leveraging the memory capabilities of LSTM networks, the controller can learn from past experiences and adapt to changing system dynamics, leading to improved control performance and stability over time.

• **Integral Terminal Sliding Mode Surface**: AITS control within the AITSM controller ensures finite-time convergence of the system. This capability allows the exoskeleton to reach its target state within a predefined timeframe, vital in time-sensitive applications.

Numerical simulations confirm the effectiveness of the proposed control approach, demonstrating its superiority over traditional methods. It's also worth noting that this method has been benchmarked against methodologies presented in other research papers such as [30].

## 2. System Description

In this section, we explore the dynamic model of a 5-DOF upper-limb exoskeleton robot, specifically designed for zero-force control. The model is explained using Lagrange and state space equations. Equation (1) outlines the Lagrange equation, which forms a connection between the robot's position, velocity, and acceleration vectors with its torque inputs, external forces, and elements such as inertia, Coriolis effects, and gravity.

$$\boldsymbol{M}(\boldsymbol{q})\ddot{\boldsymbol{q}} + \boldsymbol{C}(\boldsymbol{q},\dot{\boldsymbol{q}})\dot{\boldsymbol{q}} + \boldsymbol{G}(\boldsymbol{q}) + \boldsymbol{F} = \boldsymbol{u}, \tag{1}$$

where $\boldsymbol{q}, \dot{\boldsymbol{q}}, \ddot{\boldsymbol{q}} \in \boldsymbol{R}^{5}$, all belonging to $\boldsymbol{R}^{5}$, represent the position, velocity, and acceleration vectors of the robot, respectively. The variables $\boldsymbol{F}$ and $\boldsymbol{u}$ denote the external forces and torque inputs, respectively. $\boldsymbol{M}(\boldsymbol{q})$, $\boldsymbol{C}(\boldsymbol{q},\dot{\boldsymbol{q}})$, and $\boldsymbol{G}(\boldsymbol{q})$ represent the inertia matrix, Coriolis, and centripetal matrix, respectively. The system's nominal values are referenced in [31].

In this scenario, the vectors for position, velocity, and acceleration of the robot are represented by $\boldsymbol{q}, \dot{\boldsymbol{q}}, \ddot{\boldsymbol{q}} \in \boldsymbol{R}^{5}$ respectively, all of which belong to $\boldsymbol{R}^{5}$. The external forces and torque inputs are denoted by the variables $\boldsymbol{F}$ and $\boldsymbol{u}$, respectively. The inertia matrix, Coriolis, and centripetal matrix are represented by $\boldsymbol{M}(\boldsymbol{q})$, $\boldsymbol{C}(\boldsymbol{q},\dot{\boldsymbol{q}})$, and $\boldsymbol{G}(\boldsymbol{q})$, respectively. The nominal values of the system are referred to in the work of [31].

- **Assumption 1:** All vectors that denote position, velocity, and acceleration are quantifiable and have well-established upper bounds. Furthermore, their first and second derivatives are not only existent but also remain within limits over time.
- **Assumption 2:** The inertia matrix $\boldsymbol{M}(\boldsymbol{q})$ is both symmetric and positive definite, with a well-defined upper boundary.
- **Assumption 3:** The input force vector $\boldsymbol{F}$ is constrained by a constant $\boldsymbol{K}$, ensuring that the norm of $\| \boldsymbol{F} \|$ does not surpass $\boldsymbol{K}$, represented as $\| \boldsymbol{F} \| \leq \boldsymbol{K}$. This limitation on $\boldsymbol{K}$ imposed to maintain stability and precision within the control system.

Zero-force control is a technique employed in robotics and automation systems to attain a specific force or torque at the end effector, like a gripper or tool, of a robot. This is in contrast to position control, which aims at achieving certain positions or paths. The Denavit-Hartenberg parameters, which are vital for zero-force control, are elaborated in [30]. The transformation matrix, which maps the origin of the fixed member to the hand, can be derived using the following equation.

$$T = T_0^1 \times T_1^2 \times T_2^3 \times T_3^4 \times T_4^5, \tag{2}$$

The computation of the transformation matrix between the two members is carried out using the Denavit-Hartenberg parameters. The specific calculation method is as follows:

$$T_0^1 = \begin{bmatrix} cos(q_0) & 0 & sin(q_0) & 60.35\, sin(q_0) \\ sin(q_0) & 0 & -cos(q_0) & 60.35\, cos(q_0) \\ 0 & 1 & 0 & 0 \\ 0 & 0 & 0 & 1 \end{bmatrix}, \tag{3}$$

$$T_1^2 = \begin{bmatrix} cos(q_1) & -sin(q_1) & 0 & scap\, cos(q_1) \\ sin(q_1) & cos(q_1) & 0 & scap\, sin(q_1) \\ 0 & 0 & 1 & 0 \\ 0 & 0 & 0 & 1 \end{bmatrix}, \tag{4}$$

$$T_2^3 = \begin{bmatrix} sin(q_2) & 0 & cos(q_2) & l\,i\,nk\,cos(q_2) \\ -cos(q_2) & 0 & sin(q_2) & l\,i\,nk\,sin(q_2) \\ 0 & -1 & 0 & 0 \\ 0 & 0 & 0 & 1 \end{bmatrix}, \tag{5}$$

$$T_3^4 = \begin{bmatrix} 0 & -sin(q_3) & -cos(q_3) & -cos(q_3) \\ 0 & cos(q_3) & -sin(q_3) & -sin(q_3) \\ 1 & 0 & 0 & 0 \\ 0 & 0 & 0 & 1 \end{bmatrix}, \tag{6}$$

$$T_4^5 = \begin{bmatrix} cos(q_4) & -sin(q_4) & 0 & 0 \\ sin(q_4) & cos(q_4) & 0 & 0 \\ 0 & 0 & 1 & -fullarm \\ 0 & 0 & 0 & 1 \end{bmatrix}. \tag{7}$$

By applying the transformation matrix derived from equations (2) to (7), we can easily ascertain the motion of each joint in relation to the movements of the final actuator. By taking the first three rows from the last column of this matrix, we acquire a vector that signifies the position of the final actuator, represented as $\boldsymbol{x_e}$. Based on this information, we can then formulate the Jacobian matrix.

$$\dot{\boldsymbol{x}}_e = \boldsymbol{J}\dot{\boldsymbol{q}}, \tag{8}$$

According to (8), by employing the derived Jacobian matrix from equation (8), we can determine the resultant torque in each joint, which is induced by the force exerted on the final actuator.

$$\boldsymbol{F} = \boldsymbol{J}^T\boldsymbol{F}_e, \tag{9}$$

here, $\boldsymbol{F}$ represents the force that is input into the Zero-force control, and $\boldsymbol{F}_e$ is the three-dimensional force.

## 3. Controller Description

This section introduces the equations and architecture of the ITSM controller, PPO system, and the proposed AITSM controller integrated with PPO. These elements are crucial to the paper as they provide a mathematical basis for the proposed controller.

### 3.1 Non-Adaptive Sliding Mode Controller

The subsequent equations establish the desired output and tracking error for the system. The desired output is characterized as a state vector, represented by

$$\boldsymbol{E} = \boldsymbol{F_d} - \boldsymbol{F}\,, \qquad \boldsymbol{E} = [E_1, E_2, E_3, E_4, E_5]^T, \tag{10}$$

Here, $\boldsymbol{E}$ represents the force tracking error, and $\boldsymbol{F_d}$ is a zero vector, given that the target force control is zero. To implement zero-force control, the robot's dynamic equation in (1) is converted into the form presented in equation (12).

$$\boldsymbol{M}(\tilde{\boldsymbol{q}})\ddot{\tilde{\boldsymbol{q}}} + \boldsymbol{C}(\tilde{\boldsymbol{q}}, \dot{\tilde{\boldsymbol{q}}})\dot{\tilde{\boldsymbol{q}}} + \boldsymbol{G}(\tilde{\boldsymbol{q}}) = \boldsymbol{u} - \boldsymbol{E}, \tag{11}$$

For simplicity, the following notation has been chosen.

$$\boldsymbol{G}(\tilde{\boldsymbol{q}}, \dot{\tilde{\boldsymbol{q}}}, \ddot{\tilde{\boldsymbol{q}}}) = \boldsymbol{M}(\tilde{\boldsymbol{q}})\ddot{\tilde{\boldsymbol{q}}} + \boldsymbol{C}(\tilde{\boldsymbol{q}}, \dot{\tilde{\boldsymbol{q}}})\dot{\tilde{\boldsymbol{q}}} + \boldsymbol{G}(\tilde{\boldsymbol{q}}) \tag{12}$$

The traditional integral sliding surface for this system is characterized in (13).

$$\boldsymbol{s} = \boldsymbol{E} + \alpha_1 \int_0^T \boldsymbol{E}\, dt \qquad\qquad \alpha_1 > 0, \tag{13}$$

The terminal integral sliding surface, as proposed, is defined as follows.

$$\boldsymbol{s} = \boldsymbol{E} + \alpha_1 \int_0^T \boldsymbol{E}\, dt + \alpha_2 \int_0^T |\boldsymbol{E}|^{\gamma} sign(\boldsymbol{E}) dt \qquad\qquad \alpha_2, \alpha_1 > 0, \qquad 0 < \gamma < 2 \tag{14}$$

**Remark 1.** The proposed terminal integral sliding surface's ability to achieve faster convergence and improved robustness over the conventional integral sliding surface can lead to enhanced system performance. This includes better tracking accuracy, reduced settling time, and improved disturbance rejection capabilities.

**Remark 2.** The inclusion of the exponential term $|\boldsymbol{E}|^{\gamma} sign(\boldsymbol{E})$ in the proposed terminal integral sliding surface helps to accelerate the system's response time. This is particularly beneficial in applications where rapid system response is critical.

By computing the derivative of the proposed terminal integral sliding surface, we obtain

$$\dot{\boldsymbol{s}} = \dot{\boldsymbol{E}} + \alpha_1 \boldsymbol{E} + \alpha_2 |\boldsymbol{E}|^{\gamma} sign(\boldsymbol{E}) \qquad \alpha_1 > 0, \tag{15}$$

The control signal is composed of two elements: the equivalent term, denoted as $u_{eq}$, and the switching term, represented as $u_{sw}$. The design of the equivalent term is as follows.

$$\dot{\boldsymbol{s}} = \dot{\boldsymbol{E}} + \alpha_1 \boldsymbol{E} + \alpha_2 |\boldsymbol{E}|^{\gamma} sign(\boldsymbol{E}) = 0 \tag{16}$$

Using (14) and (16), we obtain

$$\dot{\boldsymbol{G}}(\tilde{\boldsymbol{q}}, \dot{\tilde{\boldsymbol{q}}}, \ddot{\tilde{\boldsymbol{q}}}) - \dot{\boldsymbol{u}}_{\boldsymbol{eq}} + \alpha_1 \boldsymbol{E} + \alpha_2 |\boldsymbol{E}|^{\gamma} sign(\boldsymbol{E}) = 0 \tag{17}$$

The equivalent control term, denoted as $\boldsymbol{u}_{\boldsymbol{eq}}$, and its derivative, represented as $\dot{\boldsymbol{u}}_{\boldsymbol{eq}}$, are derived as follows:

$$\dot{\boldsymbol{u}}_{\boldsymbol{eq}} = \dot{\boldsymbol{G}}(\tilde{\boldsymbol{q}}, \dot{\tilde{\boldsymbol{q}}}, \ddot{\tilde{\boldsymbol{q}}}) + \alpha_1 \boldsymbol{E} + \alpha_2 |\boldsymbol{E}|^{\gamma} sign(\boldsymbol{E}) \tag{18}$$

$$\boldsymbol{u}_{\boldsymbol{eq}} = \boldsymbol{G}(\tilde{\boldsymbol{q}}, \dot{\tilde{\boldsymbol{q}}}, \ddot{\tilde{\boldsymbol{q}}}) + \alpha_1 \int_0^T \boldsymbol{E}\, dt + \alpha_2 \int_0^T |\boldsymbol{E}|^{\gamma} sign(\boldsymbol{E}) dt \tag{19}$$

The control signal of the suggested ITSM controller is composed of two components, namely $u_{eq}(t)$ and $u_{sw}(t)$. Therefore, the control signal can be represented as follows:

$$\boldsymbol{u} = \boldsymbol{u}_{\boldsymbol{eq}} + \boldsymbol{u}_{\boldsymbol{sw}} \tag{20}$$

In order to accelerate the convergence rate and reduce the chattering effect, a new exponential switching term (reaching law) is formulated in equation (21).

$$\boldsymbol{u}_{\boldsymbol{sw}} = K_1 sign(\boldsymbol{s}) + K_2 |\boldsymbol{s}|^{0.5} sign(\boldsymbol{s}) + \left({K_3}^{|\boldsymbol{s}|} - 1\right) \tag{21}$$

The final control signal is calculated as follows:

$$\boldsymbol{u} = \boldsymbol{G}(\tilde{\boldsymbol{q}}, \dot{\tilde{\boldsymbol{q}}}, \ddot{\tilde{\boldsymbol{q}}}) + \alpha_1 \int_0^T \boldsymbol{E}\, dt + \alpha_2 \int_0^T |\boldsymbol{E}|^{\gamma} sign(\boldsymbol{E}) dt + K_1 sign(\boldsymbol{s}) + K_2 |\boldsymbol{s}|^{0.5} sign(\boldsymbol{s}) + \left({K_3}^{|\boldsymbol{s}|} - 1\right) \tag{22}$$

**Remark 3:** The proposed control law incorporates two key terms to achieve adaptive convergence and mitigate chattering. When the system states deviate significantly from the desired trajectory, the term $\left(K_3{}^{|\boldsymbol{s}|}-1\right)$ helps to decelerate convergence, while near or on the sliding surface, it becomes insignificant. The exponential term accelerates convergence towards the sliding surface, and the term $K_2|\boldsymbol{s}|^{0.5}sign(\boldsymbol{s})$ ensures a smooth and continuous control signal, mitigating chattering in the vicinity of the sliding surface. This integrated approach enables rapid convergence, minimal chattering, and system stability.

Theorem 1 is introduced to illustrate the Lyapunov stability of the system when governed by the proposed ITSM controller.

**Theorem 1:** Given that the dynamic system (1) satisfies assumptions (1-3) and is controlled by the proposed control signal (22), the tracking error of the system will approach zero within a specified time frame. Furthermore, the closed-loop stability of the system, when operated by the ITSM controller, is guaranteed.

**Proof.**

Equation (23) formulates a Lyapunov candidate function which symbolizes the tracking error of the system.

$$V_1=\frac{1}{2}\boldsymbol{s}^2 \tag{23}$$

The derivative with respect to time of the proposed Lyapunov function is derived as follows.

$$\dot{V}_1=\boldsymbol{s}\dot{\boldsymbol{s}} \tag{24}$$

Equation (24) can be reshaped into the structure of (25), by applying the principles defined in (16), thereby clarifying the dynamics of the error system.

$$\dot{V}_1=\boldsymbol{s}\left(\dot{\boldsymbol{E}}+\alpha_1\boldsymbol{E}+\ \alpha_2|\boldsymbol{E}|^{\gamma}sign(\boldsymbol{E})\right) \tag{25}$$

with simplification of (25), yields

$$\dot{V}_1 = -\boldsymbol{s}\left(\left(\left(K_3{}^{|\boldsymbol{s}|} - 1\right) + K_1\right) sign(\boldsymbol{s}) + K_2|\boldsymbol{s}|^{0.5} sign(\boldsymbol{s})\right) \tag{26}$$

Equation (26) demonstrates that the derivative of $V_1$, represented as $\dot{V}_1$, is semi-negative definite, indicating that its value is consistently non-positive.

To prove the stability of the system, it is necessary to show that the expression enclosed in parentheses is negative for $\boldsymbol{s} > 0$ and positive for $\boldsymbol{s} < 0$. This is achieved by leveraging the facts that:

- $\boldsymbol{s}$ and $sign(\boldsymbol{s})$ have the same sign when $\boldsymbol{s} > 0$ and opposite signs when $\boldsymbol{s} < 0$.
- $K_1$, $K_2$, and $K_3$ are positive constants

As a result, the following inequalities hold:

$$\left(\left(K_3{}^{|\boldsymbol{s}|} - 1\right) + K_1\right) sign(\boldsymbol{s}) + K_2|\boldsymbol{s}|^{0.5} sign(\boldsymbol{s}) < 0, \qquad for\ \ \boldsymbol{s} > 0 \tag{27}$$

$$\left(\left(K_3{}^{|\boldsymbol{s}|} - 1\right) + K_1\right) sign(\boldsymbol{s}) + K_2|\boldsymbol{s}|^{0.5} sign(\boldsymbol{s}) > 0, \qquad for\ \ \boldsymbol{s} < 0 \tag{28}$$

Therefore, in both scenarios, it has been demonstrated that $\dot{V}_1$ is negative definite. To proceed with the proof using LaSalle's Invariance Principle, we must identify the largest invariant set where $\dot{V}_1 = 0$ and establish that the only solution within this set is the equilibrium point at $\boldsymbol{s} = \boldsymbol{0}$. It is evident that $\dot{V}_1 = 0$ if and only if

$$\left(\left(K_3{}^{|\boldsymbol{s}|} - 1\right) + K_1\right) sign(\boldsymbol{s}) + K_2|\boldsymbol{s}|^{0.5} sign(\boldsymbol{s}) = 0 \tag{28}$$

This implies that

$$\left(\left(K_3{}^{|\boldsymbol{s}|} - 1\right) + K_1\right) = -K_2|\boldsymbol{s}|^{0.5} \tag{29}$$

By applying the absolute value to both sides and reorganizing, we obtain

$$\left(K_3{}^{|\boldsymbol{s}|} - 1\right) = -K_2|\boldsymbol{s}|^{0.5} - K_1 \tag{30}$$

Given that the left-hand side is non-negative, it is necessary that

$$-K_2|\boldsymbol{s}|^{0.5} - K_1 \geq 0 \tag{31}$$

or

$$|\boldsymbol{s}| \geq \left({}^{K_1}/_{K_2}\right)^2 \tag{32}$$

However, this is in conflict with the fact that $V_1$ is positive definite and radially unbounded, which suggests that the absolute value of s must be limited. Consequently, the only feasible solution that satisfies $\dot{V}_1 = 0$ is when $\boldsymbol{s}$ equals 0. By invoking LaSalle's Invariance Principle, it is inferred that every solution that originates in a vicinity of $\boldsymbol{s} = \boldsymbol{0}$ converges to $\boldsymbol{s} = \boldsymbol{0}$ as $t \to \infty$. Hence, $\boldsymbol{s} = \boldsymbol{0}$ is asymptotically stable. This concludes the proof.

## 3.2 Adaptive Sliding Mode Controller using PPO

PPO is a type DRL algorithm that enables an agent to learn decision-making by interacting with an environment and receiving rewards or penalties. PPO updates the policy in a controlled manner using a clipping parameter, ensuring stable and efficient learning. The ultimate goal of PPO is to find the optimal policy that maximizes the expected cumulative reward, which is formulated as an optimization problem.

$$J(\theta) = \mathbb{E}_t\left[\frac{\pi_\theta(a_t|s_t)}{\pi_{\theta_{old}}(a_t|s_t)}A_t\right], \tag{33}$$

here, $\pi_\theta(a_t|s_t)$ signifies the current policy, while $\pi_{\theta_{old}}(a_t|s_t)$ stands for the old policy. The old policy acts as a benchmark to ensure that the updates to the policy do not stray too far from the previously learned policy, thereby aiding in maintaining stability during the training process. $A_t$ is the advantage function, which quantifies the relative benefit of an action compared to the average action for that particular state. The following formula shows the advantage function.

$$A_t = Q(s_t, a_t) - V(s_t), \tag{34}$$

$Q(s_t, a_t)$ is the action-value function, which represents the expected return (cumulative reward) of taking action $a_t$ in state $s_t$ and following the policy thereafter. $V(s_t)$ is the value function, which represents the expected return of being in state $s_t$ and following the policy thereafter. To prevent large policy updates, PPO introduces a clipping term. In order to avoid substantial updates to the policy, PPO algorithm incorporates a clipping term. This results in a modification of the objective function. The objective is modified in (35).

$$J^{\text{CLIP}}(\theta) = \mathbb{E}_t\left[min\left(\frac{\pi_\theta(a_t|s_t)}{\pi_{\theta_{old}}(a_t|s_t)}A_t, \text{clip}\left(1-\varepsilon, 1+\varepsilon, \frac{\pi_\theta(a_t|s_t)}{\pi_{\theta_{old}}(a_t|s_t)}\right)A_t\right)\right], \tag{35}$$

here $\varepsilon$ represents the clipping parameter, and the function $\text{clip}(x, y, z)$ restricts the value of $z$ to lie within the range $[x, y]$. The clipping parameter in PPO is a hyperparameter that dictates the magnitude of policy updates. It’s generally chosen to be a value between 0.1 and 0.5. A larger value of $\varepsilon$ leads to bigger policy updates, but it may also cause the learning process to become more unstable. Conversely, a smaller ε value results in smaller policy updates, contributing to a more stable learning process. The policy parameters, denoted by $\theta$, are updated by optimizing the clipped surrogate objective.

$$\theta_{new} = \arg\max_\theta J^{\text{CLIP}}(\theta)\,. \tag{36}$$

This update rule guarantees that the policy doesn't stray significantly from the previous policy in a single iteration, thereby ensuring a stable learning process during training.

In this problem, The PPO algorithm is being adapted for a control problem where the objective is to determine the optimal control gains ($K_1$, $K_2$, and $K_3$). These gains are part of a reaching law in SMC, which is a robust control strategy used to drive the system's states to a predefined sliding surface and maintain them there despite disturbances. The proposed objective function is given in (37).

$$J_{PPO-L}(\theta) = \mathbb{E}_t\left[\min\left(\frac{\pi_\theta(a_t|s_t)}{\pi_{\theta_{\text{old}}}(a_t|s_t)}A_t, \text{clip}(1-\epsilon, 1+\epsilon, \frac{\pi_\theta(a_t|s_t)}{\pi_{\theta_{\text{old}}}(a_t|s_t)})A_t\right) - \beta(V_1(s) + V_2(s))\right], \quad (37)$$

The modified PPO objective function, denoted as $J_{PPO-L}$, integrates the Lyapunov function where $V_1(s) = \frac{1}{2}\boldsymbol{s}^2$ representing the tracking error of the system and $V_1(s) = \lambda \times |\boldsymbol{s}|$ is an additional term that penalizes the magnitude of the state, where $\lambda$ is a hyperparameter.

This term penalizes the growth of the Lyapunov function $V_1(\boldsymbol{s})$ and the magnitude of the state $\boldsymbol{s}$, encouraging the policy to take actions that reduce the tracking error and maintain stability. By incorporating the Lyapunov function and the additional penalty term into the PPO objective function, we can encourage the policy to optimize the control gains for the SMC problem, while maintaining stability and reducing the tracking error.

## 3.3 The proposed Deep Neural Network

The proposed deep neural network $NN_\theta$ takes the state $s_t$ as input and outputs the control gains ($K_1$, $K_2$, and $K_3$). The neural network consists of a LSTM layer with an attention mechanism. The attention mechanism enables the model to selectively focus on specific parts of the system's state sequence, allowing it to weigh the importance of different elements and compute a context vector that captures the most relevant information. This context vector is then used to output the adapted control gains. The attention mechanism's ability to dynamically adjust its focus on different parts of the input sequence makes it particularly well-suited for control problems, where the system's behavior can change over time. This novel technique offers several traits that are beneficial in control problems, including the ability to handle sequential data, adapt to changing system dynamics, and provide interpretable results through the attention weights. Additionally, the use of an LSTM layer allows the model to capture long-term dependencies in the system's behavior, further improving its ability to adapt the control gains effectively. Overall, this approach provides a powerful tool for adaptive control, enabling the development of more efficient and effective control systems.

In LSTM, the input sequence is $s_t$, which represents the system's state at time $t$. The LSTM layer consists of three components:

- Cell State: The cell state $c_t$ represents the internal memory of the LSTM layer, which captures information from previous time steps.
- Hidden State: The hidden state $h_t$ represents the output of the LSTM layer at time $t$, which is used to compute the attention weights.
- Gates: The gates are learnable parameters that control the flow of information into and out of the cell state and hidden state.

The LSTM layer can be formalized as follows:

$h_t, c_t = LSTM(s_t, h_{t-1}, c_{t-1})$

where $h_t$ and $c_t$ are the hidden state and cell state, respectively, and $h_{t-1}$ and $c_{t-1}$ are the previous hidden state and cell state, respectively.

The LSTM layer computes the following gates:

- Input Gate: $i_t = \sigma\ (W_i \times s_t + U_i \times h_{t-1} + b_i)$ (38)
- Forget Gate: $f_t = \sigma\left(W_f \times s_t + U_f \times h_{t-1} + b_f\right)$ (39)
- Output Gate: $o_t = \sigma\ (W_o \times s_t + U_o \times h_{t-1} + b_o)$ (40)
- Cell State Update: $c_t = f_t \times c_{t-1} + i_t \times Tanh\ (W_c \times s_t + U_c \times h_{t-1} + b_c)$ (41)
- Hidden State Update: $h_t = o_t \times Tanh(c_t)$ (42)

where $\sigma$ is the sigmoid function, $Tanh$ is the hyperbolic tangent function, and $W_i$, $U_i$, $b_i$, $W_f$, $U_f$, $b_f$, $W_o$, $U_o$, $b_o$, $W_c$, $U_c$, and $b_c$ are learnable parameters.

The attention mechanism computes a weighted sum of the hidden states $h_t$ to obtain a context vector $c_t$. The attention weights $\alpha_t$ are computed using the hidden states $h_t$ and learnable parameters $W_a$ and $b_a$. The attention mechanism can be formalized as follows:

$$\alpha_t = Softmax(W_a \times h_t + b_a) \tag{43}$$

where $c_t$ is the attention weight at time $t$ , and $W_a$ and $b_a$ are learnable parameters. The context vector $c_t$ is computed as a weighted sum of the hidden states $h_t$ using the attention weights $\alpha_t$:

$$c_t = \sum \alpha_t \times h_t \tag{44}$$

where $c_t$ is the context vector at time $t$ .

The attention mechanism allows the model to focus on specific parts of the input sequence when computing the context vector $c_t$. This is particularly useful in sequential data processing tasks, such as natural language processing and time series forecasting.

The gains ($K_1$, $K_2$, and $K_3$) are part of the action $a_t$ and are determined by the neural network $NN_\theta$ based on the current state $s_t$. The neural network is designed to output these gains, which are then used to calculate the control input according to a specific reaching law.

$$K_1(s_t) = NN_\theta^{K_1}(s_t) \tag{45}$$

$$K_2(s_t) = NN_\theta^{K_2}(s_t) \tag{46}$$

$$K_3(s_t) = NN_\theta^{K_3}(s_t) \tag{47}$$

Hence; the proposed control signal has been changed to (41) in the adaptive case.

$$\boldsymbol{u} = \boldsymbol{G}(\tilde{\boldsymbol{q}}, \dot{\tilde{\boldsymbol{q}}}, \ddot{\tilde{\boldsymbol{q}}}) + \alpha_1 \int_0^T \boldsymbol{E}\, dt + \alpha_2 \int_0^T |\boldsymbol{E}|^\gamma sign(\boldsymbol{E}) dt + \left(NN_\theta^{K_1}(s_t)\right) sign(\boldsymbol{s}) + \left(NN_\theta^{K_2}(s_t)\right)|\boldsymbol{s}|^{0.5} sign(\boldsymbol{s}) + \left(\left(NN_\theta^{K_3}(s_t)\right)^{|\boldsymbol{s}|} - 1\right) \tag{48}$$

By optimizing this objective function, the PPO algorithm learns a policy that not only maximizes the expected cumulative reward (reflecting the control performance) but also ensures the stability of the control system by minimizing the Lyapunov function. The block diagram of the proposed neural network using PPO is shown in Fig. 1.

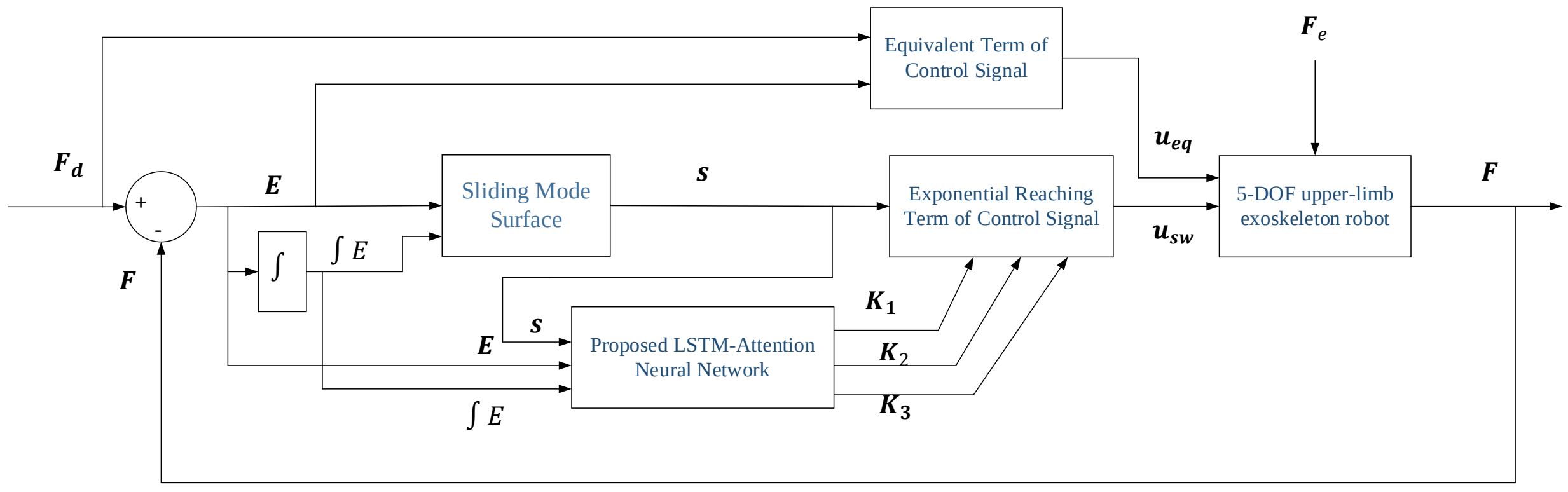

Fig. 1. The block diagram of the proposed neural network using PPO for force control of 5-DOF upper-limb exoskeleton robot.

## 4. Simulation Result

In this section, we put our proposed control strategy to the test in two distinct scenarios, demonstrating its versatility, resilience, and excellence. We carry out a comparative study of the performance of the SMC, Integral Terminal Sliding Mode Controller (ITSMC), and Adaptive Integral Terminal Sliding Mode Controller (AITSMC) with the exponential term and PPO algorithm. Additionally, we evaluate the performance of the adaptive passivity-based controller, as suggested by (Khan et al., 2016), to validate the effectiveness of our proposed approach.

Equations (42) to (45) outline the formulas for the performance metrics used to evaluate our proposed adaptive control strategy in comparison with existing methods. These metrics include the Integral Square Error (ISE), Integral Time Square Error (ITSE), Average Chattering Magnitude (ACM), and Control Energy (CE). These statistical assessments play a crucial role in providing a numerical comparison of the different control strategies and in determining the effectiveness and robustness of our proposed method.

$$ISE = \int_0^T \boldsymbol{\tilde{x}}^{\boldsymbol{T}} \boldsymbol{\tilde{x}}\, dt \tag{49}$$

$$ITSE = \int_0^T t.\boldsymbol{\tilde{x}}^{\boldsymbol{T}} \boldsymbol{\tilde{x}}\, dt \tag{50}$$

$$ACM = RMS\left(\sqrt{\boldsymbol{\tilde{x}}^{\boldsymbol{T}} \boldsymbol{\tilde{x}}} - \sqrt{\boldsymbol{x_d}^{\boldsymbol{T}} \boldsymbol{x_d}}\right) \tag{51}$$

$$CE = \int_0^T (u^T u) dt \tag{52}$$

There are four common performance metrics for control systems: ISE (Integral of Squared Error), ITSE (Integral of Time-weighted Squared Error), ACM (Absolute Control Movement), and CE (Control Effort). Each metric prioritizes different aspects of control performance: ISE reduces overall deviations, ITSE prioritizes fast response but may lead to oscillations, ACM promotes smooth control action, and CE evaluates energy efficiency. The choice of metric depends on the specific control objectives of the application.

## 4.1 First Scenario

For the first scenario, we need to define the system's initial conditions as:

$$\boldsymbol{q}(0) = [0.4; 0.4; -0.1; -0.1; 0.1]$$

$$\dot{\boldsymbol{q}}(0) = [-0.4; 0.6; -1.5; -1.5; 1]$$

An external force disrupts the system in the first scenario as

$$\boldsymbol{F}_e(t) = [\sin(t)\mathrm{u}(\mathrm{t}-2), 0.1\cos(t)\,\mathrm{u}(\mathrm{t}-1), 0.2\sin(t)\mathrm{u}(\mathrm{t})]$$

Figures 2 and Table 1 present the simulation results for scenario 1, showcasing the proposed adaptive control strategy's effectiveness against other methods.

**Initial States**: The system's initial states are denoted by the vector $\boldsymbol{q}(0) = [0.4; 0.4; -0.1; -0.1; 0.1]$ . This vector signifies the initial values of the system variables at the starting time $t = 0$, and $\dot{\boldsymbol{q}}(0) = [-0.4; 0.6; -1.5; -1.5; 1]$ denotes the initial velocities of these variables, i.e., their rates of change at the outset.

**External Disturbance**: The external disturbance $\boldsymbol{F}_e(t) = [sin(t)u(t-2), 0.1\,cos(t)\,u(t-1), 0.2\,sin(t)u(t)]$ is a time-dependent function that represents external factors affecting the system. Here, $u(t)$ is unit step functions that switch on the corresponding sinusoidal disturbances at times $t = 2, t = 1$, and $t = 0$, respectively.

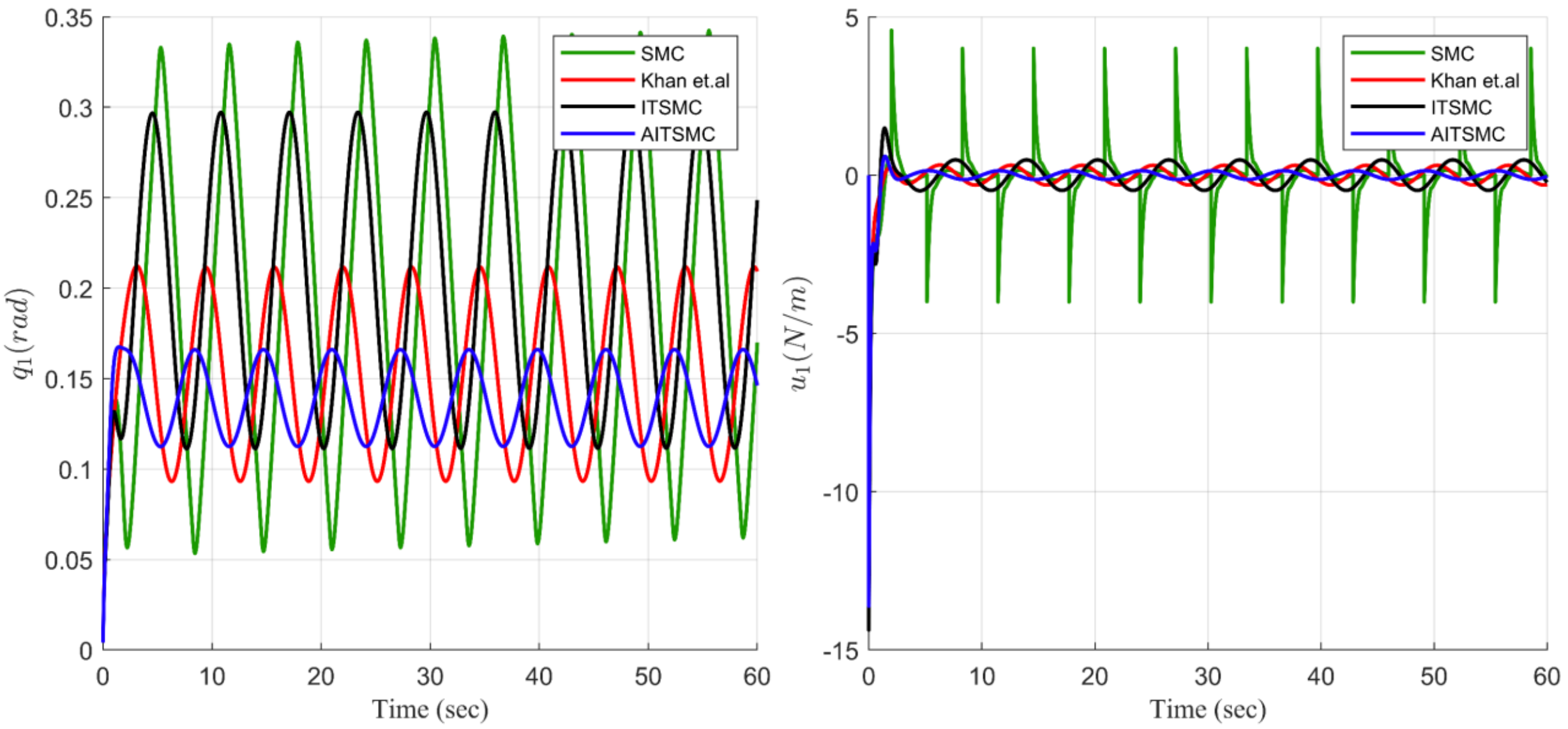

(a)

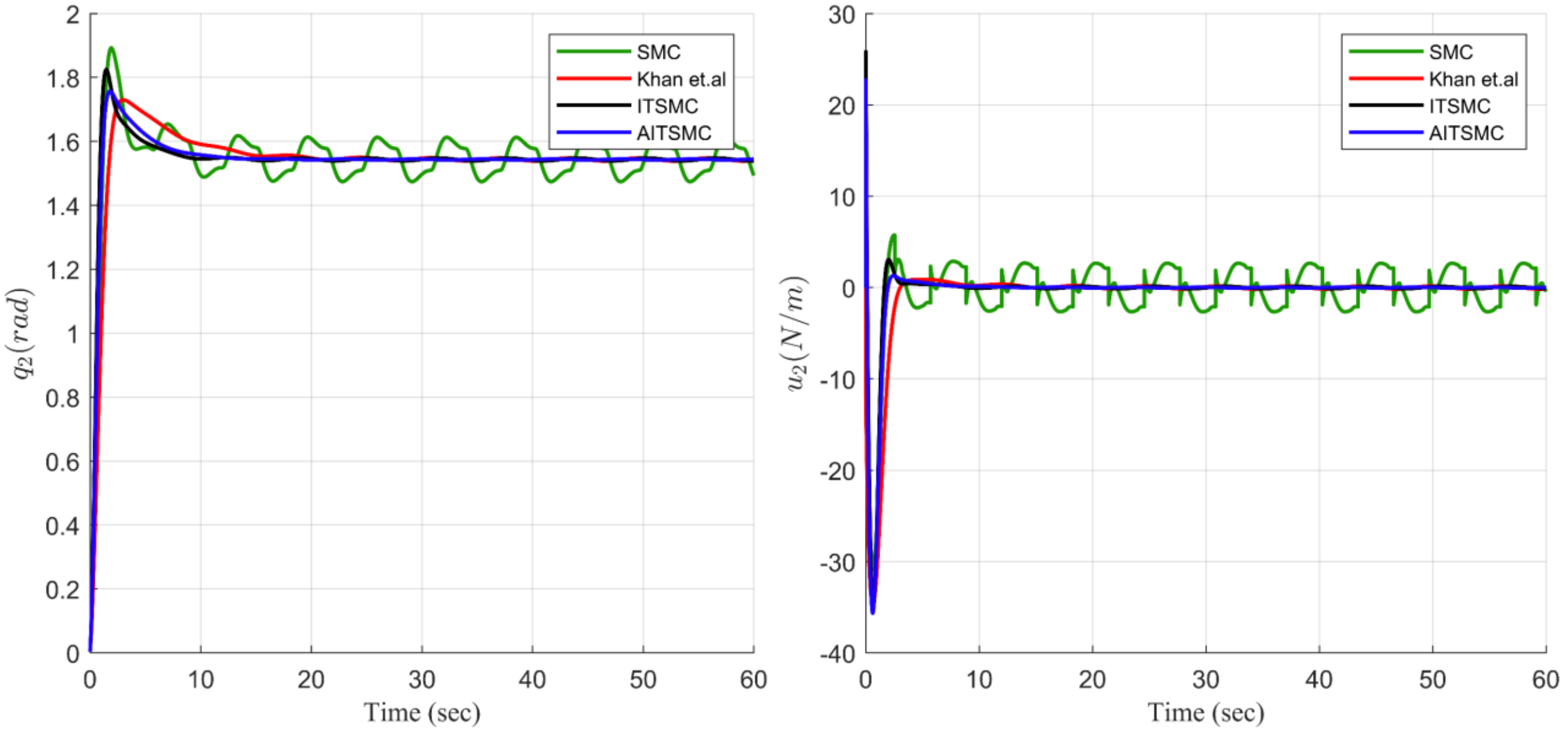

(b)

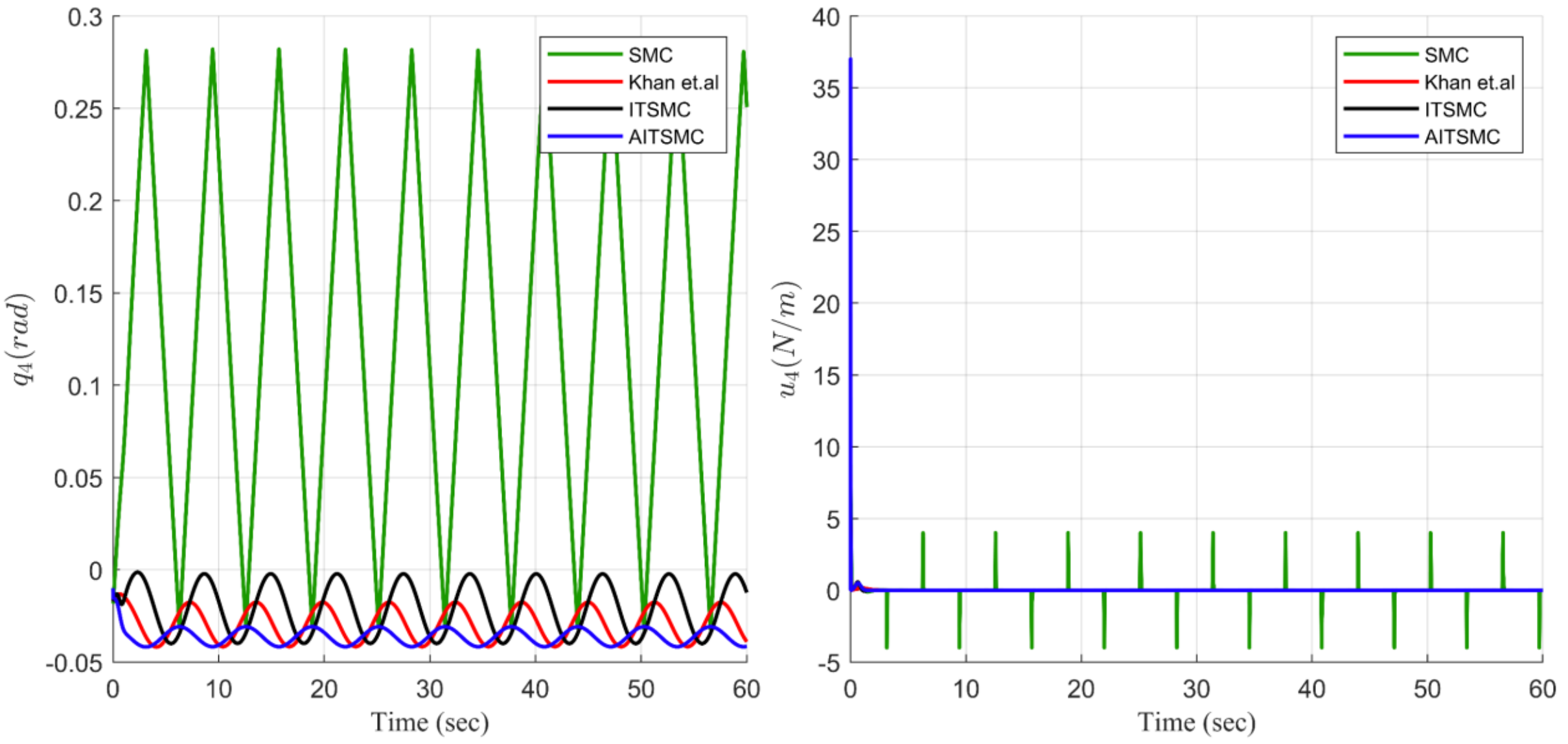

(d)

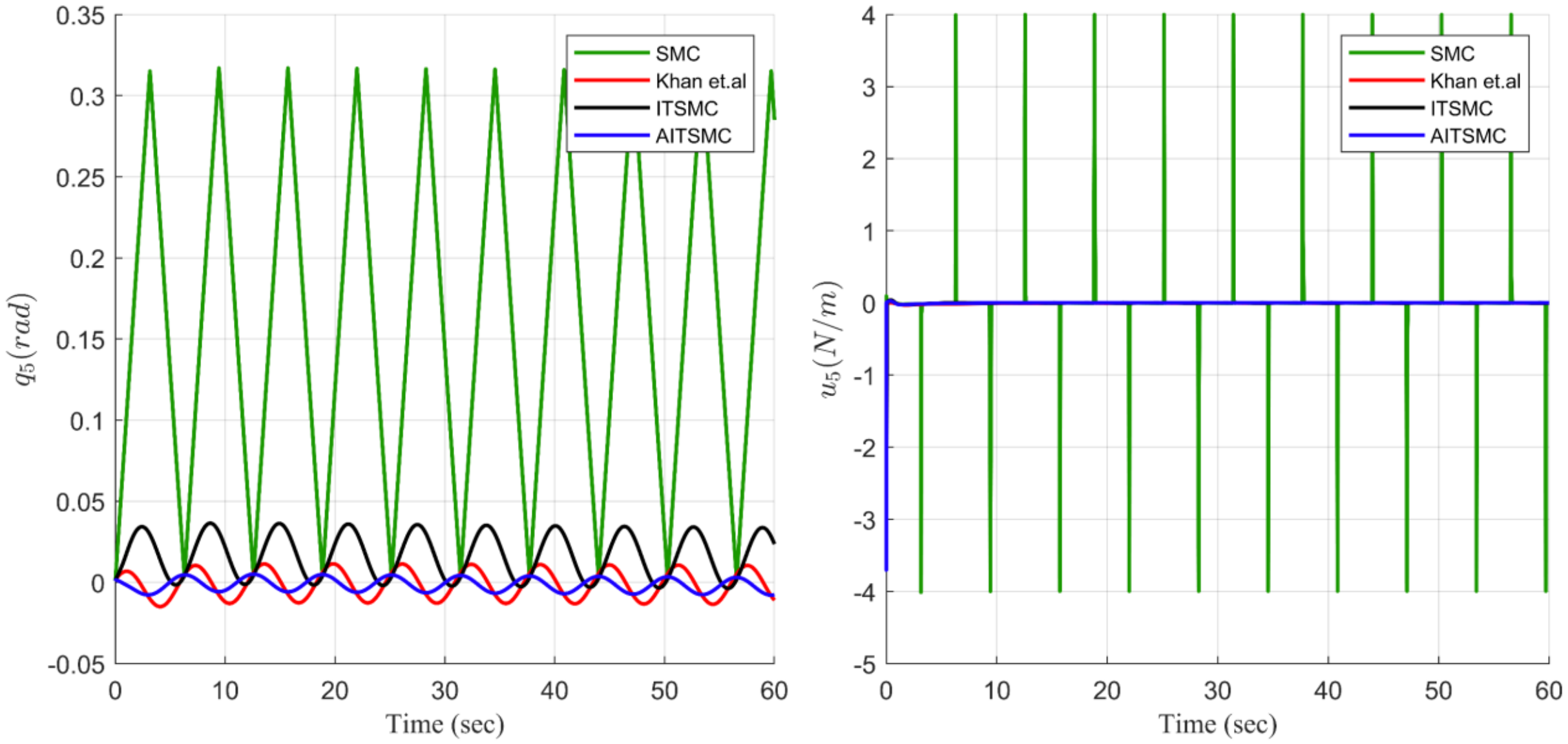

(e)

Fig. 2. The trajectory of $q$ and control signal of $u$ using the SMC, ITSMC, AITSMC with exponential term and PPO, with adaptive passivity-based [30] in the first scenario.

**Table 1.** Performance indices in the first example using the SMC, ITSMC, AITSMC with exponential term and PPO, with adaptive passivity-based [30] in the first scenario.

| **Controller** | **ISE** | **ITSE** | **CE** | **ACM** |
|---|---|---|---|---|
| SMC | **3.6057** | **4.4902** | **328947.46** | **9.7446*10$^{-2}$** |
| ITSMC with exponential term | **3.6318** | **3.3252** | **11887.85** | **7.1734*10$^{-3}$** |
| Adaptive passivity-based (Khan et al., 2016) | **1.7971** | **0.8160** | **9324.22** | **5.1365*10$^{-3}$** |
| AITSMC with exponential term | **0.9654** | **0.2343** | **6064.05** | **4.6461*10$^{-3}$** |

## 4.2 Second Scenario

For the second scenario, the initial conditions are set as follows:

- The initial states are given by $\boldsymbol{q}(0) = [-0.4, 0.8, 0.4, -1.7, 1.1]$, representing the initial values of the system variables at time $t = 0$.
- The initial rates of change of these variables are represented by $\boldsymbol{q}(0) = [-0.4, 0.8, 0.4, -1.7, 1.1]$.

The external disturbance is denoted as $\boldsymbol{F}_e(t) = [1.2\sin(t)\mathrm{u}(\mathrm{t}), 1.2\cos(t)\,\mathrm{u}(\mathrm{t}-5), -1.5\cos(t)\mathrm{u}(\mathrm{t}-4)]$. This function, which varies with time, signifies the external influences that affect the system. The effectiveness of the controllers under these conditions is evaluated using the results of numerical simulations, as shown in Figure 3 and Table 2. These results are expected to demonstrate the performance of the controllers in the face of parameter uncertainty, a crucial aspect in many real-world systems.

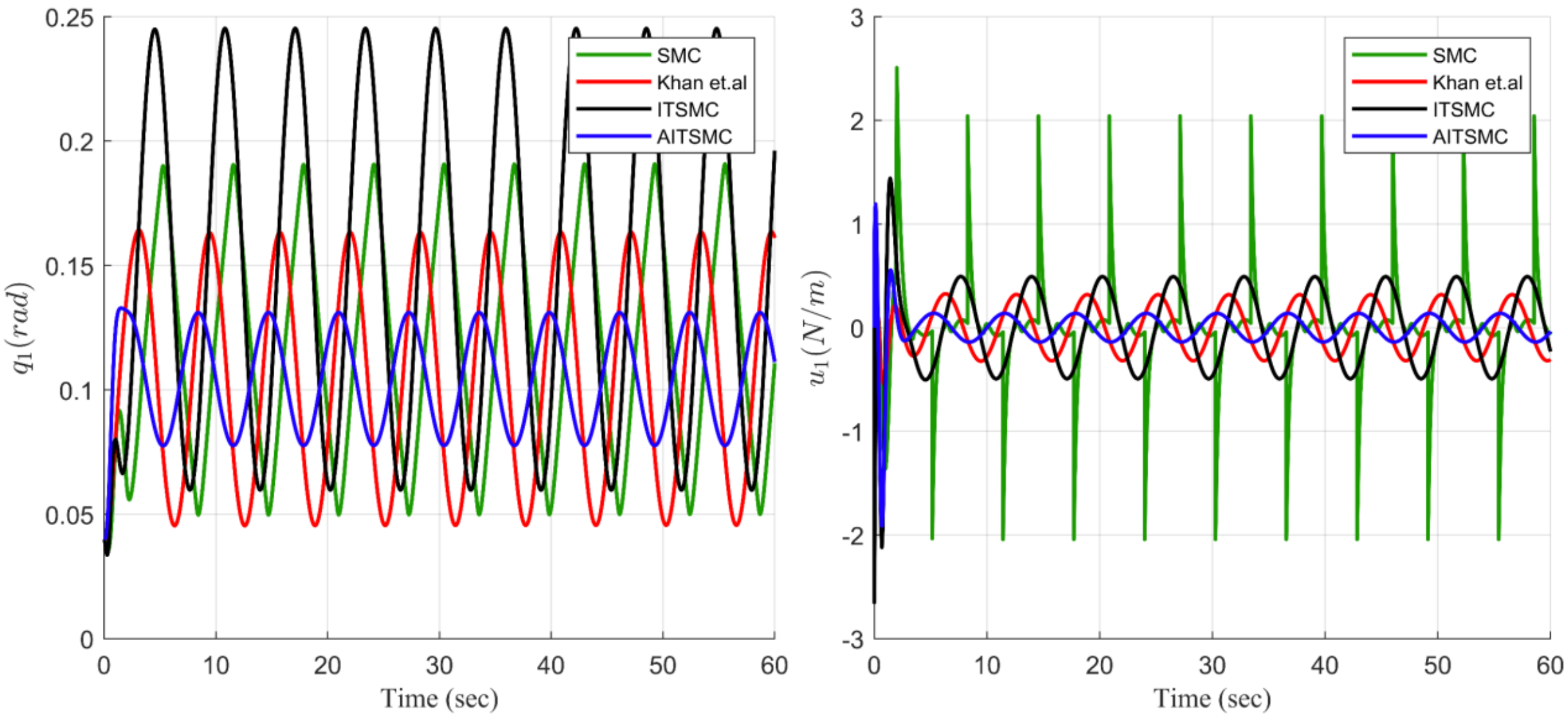

(a)

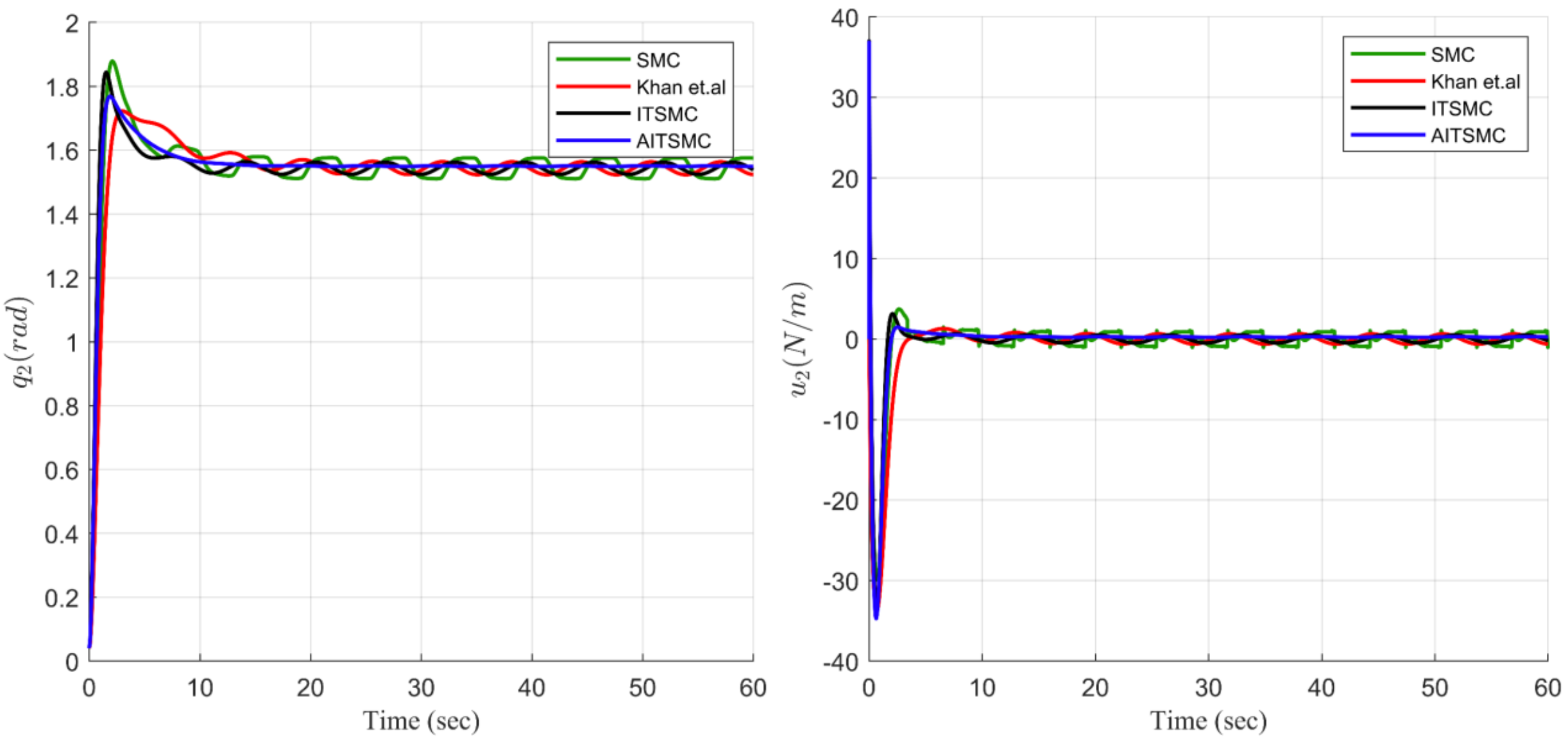

(b)

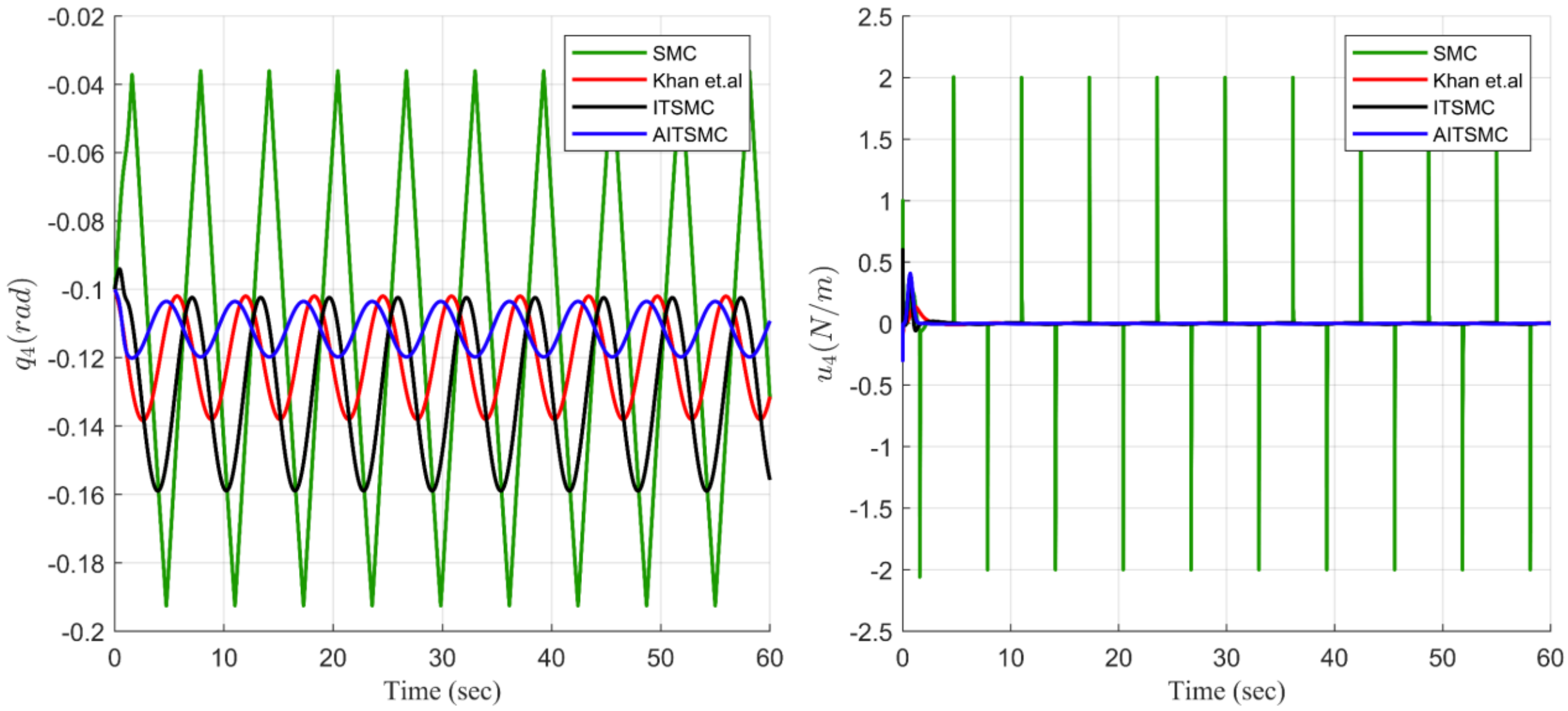

(c)

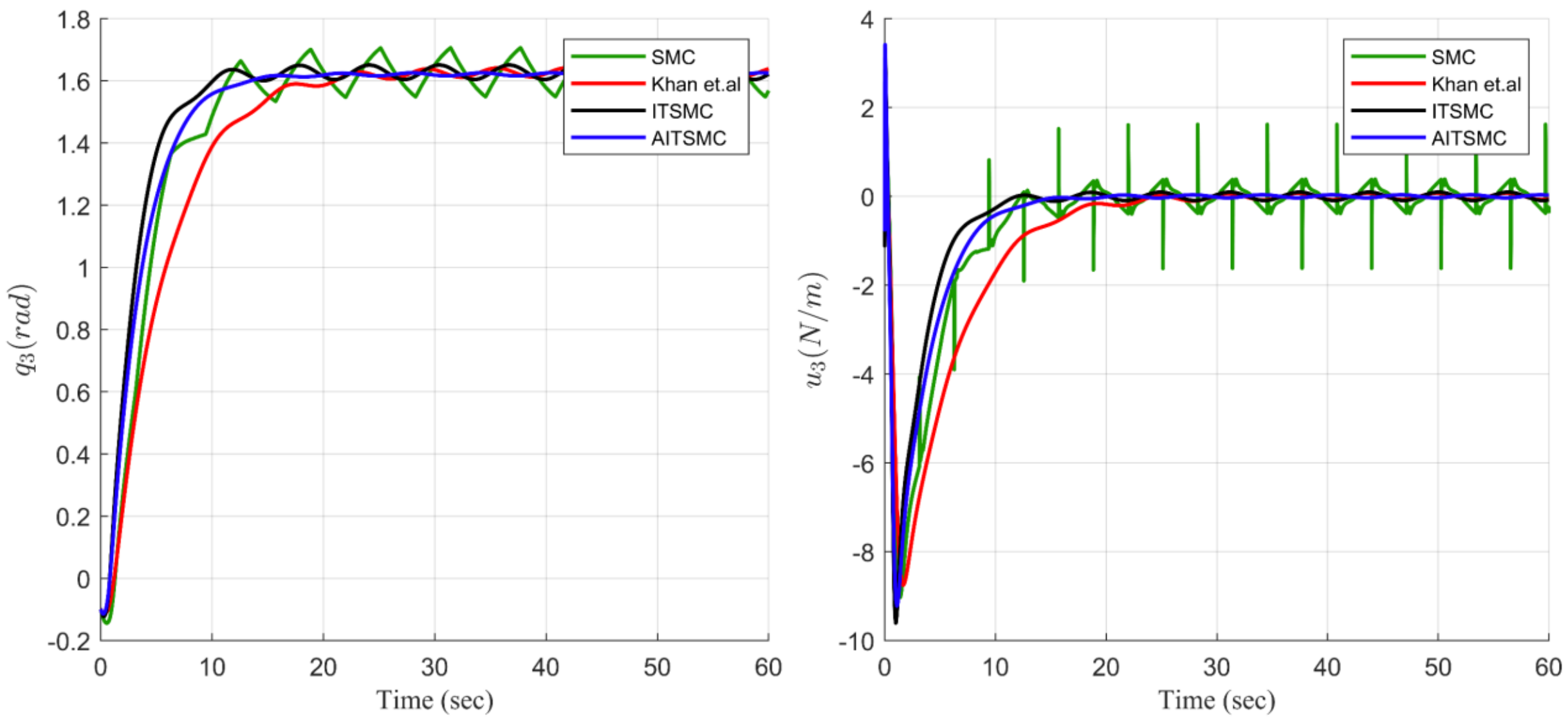

(d)

Fig. 3. The trajectory of $q$ and control signal of $u$ using the SMC, ITSMC, AITSMC with exponential term and PPO, with adaptive passivity-based [30] in the second scenario.

**Table 2.** Performance indices in the first example using the SMC, ITSMC, AITSMC with exponential term and PPO, with adaptive passivity-based (Khan et al., 2016) in the second scenario.

| Controller | ISE | ITSE | CE | ACM |
|---|---|---|---|---|
| SMC | **27.9928** | **34.9739** | **355072.23** | **2.8333*$10^{-1}$** |
| ITSMC with exponential term | **27.4683** | **24.9798** | **18014.46** | **1.2015*$10^{-1}$** |
| adaptive passivity-based (Khan et al., 2016) | **13.7801** | **6.2462** | **16452.17** | **4.6911*$10^{-2}$** |
| AITSMC with exponential term and PPO | **6.1631** | **1.4082** | **13461.64** | **4.7691 *$10^{-2}$** |

According to the Figures (2-3) and Tables (1-2), we concluded that, all controllers effectively managed and stabilized the system despite various external disturbances. They achieved a state of "zero-force control," meaning the effort required to maintain stability became minimal as the control signals converged to zero. The ITSMC stood out by significantly reducing chattering in the control signal compared to the standard SMC. Chattering can disrupt system performance, so its reduction improves overall stability. Using the exponential term in reaching law and terminal SMC, it significantly reduced chattering and increase the convergence rate of system.

The key innovation of the proposed AITSMC lies in its leveraging of the PPO algorithm, which acts as an intelligent system, continuously adjusting the controller gains to minimize the tracking error and Lyapunov function, enabling the AITSMC to effectively handle uncertainties in system dynamics and external disturbances. The AITSMC's superiority over traditional controllers is confirmed through statistical analysis, demonstrating its ability to handle various conditions and making it a more effective control strategy. Furthermore, the attention mechanism integrated into the AITSMC enables selective focus on specific parts of the system's state sequence, dynamically adjusts to changing system behavior, and offers beneficial traits such as handling sequential data, adapting to changing dynamics, providing interpretable results, and capturing long-term dependencies. The use of LSTM neural networks, well-suited for time series prediction and modeling complex sequential data, allows the attention mechanism to effectively capture and learn from the temporal relationships in the system's behavior, enabling more accurate predictions and better control decisions.

## 5. Conclusion

This study successfully introduced a novel control strategy for a 5-DOF upper-limb exoskeleton robot, leveraging the power of artificial intelligence and machine learning. The proposed approach combines the benefits of direct adaptive control, integral terminal sliding mode control, and reinforcement learning to tackle the key challenges inherent to exoskeleton systems: robustness against external force disturbances and management of parametric uncertainty. The integral terminal sliding mode surface ensures that the system reaches the desired state within a finite time, while eliminating undesirable high-frequency control signal oscillations (chattering). Moreover, the design promotes faster convergence compared to traditional methods. The attention mechanism and LSTM neural network play a crucial role in the proposed control strategy, enabling the system to selectively focus on the most relevant information and capture long-term dependencies in the system's behavior. The attention mechanism allows the system to weigh the importance of different elements and compute a context vector that captures the most relevant information, while the LSTM neural network processes the system's state sequence and outputs a hidden state that is used to compute the control gains. The PPO algorithm acts as an intelligent component, continuously adapting controller parameters to optimize performance and handle uncertainties. The effectiveness of the proposed method was confirmed through a comparative analysis with existing control strategies, demonstrating its superiority. Overall, this research lays the groundwork for further exploration and has the potential to significantly improve the control and performance of exoskeleton robots.